\pdfoutput=1
\documentclass[10pt]{article}
\usepackage[preprint]{tmlr}
\usepackage[utf8]{inputenc}
\usepackage[T1]{fontenc}
\usepackage{amsmath,amssymb,amsthm}
\usepackage{booktabs}
\usepackage{graphicx}
\usepackage{microtype}
\usepackage{xcolor}
\usepackage{hyperref}
\usepackage{url}

\graphicspath{{figures/}}
\newtheorem{proposition}{Proposition}
\newtheorem{corollary}{Corollary}
\definecolor{revblue}{RGB}{0,70,200}
\newcommand{\rev}[1]{#1}
\newcommand{\revstart}{}
\newcommand{\revend}{}
\newenvironment{revised}{}{}

\title{Who Pays for the Knowledge Models Copy?\\
Model Collapse as Lossy Transmission and Rogers' Paradox}

\author{\name Nam H. Le \email nam.le@uvm.edu \\
      \addr Department of Computer Science and Vermont Complex Systems Institute\\
      University of Vermont, Burlington, VT, USA}

\def\month{MM}
\def\year{YYYY}
\def\openreview{\url{https://openreview.net/forum?id=XXXX}}

\makeatletter
\newcommand{\coderepo}{\if@accepted at \url{https://github.com/drnamlabs-gif/rogers-model-collapse}\else
as an anonymised archive submitted as supplementary material\fi}
\makeatother

\begin{document}
\maketitle

\begin{abstract}
\revstart
Generative models trained on the output of earlier models lose knowledge
across generations, rare knowledge first. The remedies that work all bring
in information from outside the loop, such as fresh human data, verifiers,
or curators, and the theory of model collapse treats that supply as given.
We argue that cultural evolution has a theory of the supply, and that it
distinguishes two ways copying loses knowledge. In a fixed world, copies
are lossy, as in Henrich's account of skill loss in small populations. In a
changing world, even perfect copies go stale, which is Rogers' paradox. We
write both as one chain with separately measurable parameters and test it
on lineages of language models (word trigrams and a character transformer
on two books, and OPT-125m on wikitext-2, on arXiv abstracts written after
its pretraining, and on a world of 2\,000 fictional facts that change at a
set rate). In a fixed world, neutral drift on word frequencies predicts,
with no free parameter, the rare-word coverage of every trigram lineage to
within 0.03. In the fact world, the model's accuracy falls and its
outdated answers rise with the rate of change; the chain calibrated on the
static world predicts both for mid- and low-frequency facts to within 0.03
on average, with the rate of change set, not fitted. Models that inherit
their weights keep more in a static world and carry more outdated facts in
a changing one, as Rogers' theory says of individual memory. Finally we
close the loop with the people who write the data. When they can ask the
model instead of finding out, the model ends up exactly as reliable as
finding out is costly, and in both regimes someone must keep paying to
learn; we give the share in closed form, extend it to lossy copying,
heterogeneous costs, aggregating models, and users who check, and show
when checking is a stable way out.
\revend
\end{abstract}

\section{Introduction}
\label{sec:intro}

Training generative models on the output of earlier models degrades them.
Across generations, the tails of the learned distribution thin and
disappear, rare content is forgotten before common content, and the
models' output narrows \citep{shumailov2024,alemohammad2024,bertrand2024,
guo2024diversity,briesch2023}. Model collapse now has a substantial
theory: it can be described as a change of scaling laws concentrated in
the tails \citep{dohmatob2024tails}, analysed exactly in regression
\citep{dohmatob2024demystified}, shown to arise from small proportions of
synthetic data \citep{dohmatob2025strong}, bounded in rate
\citep{suresh2025,seddik2024}, and avoided when real data accumulate
rather than being replaced \citep{gerstgrasser2024,kazdan2025}. Its
definitions and assumptions are themselves debated
\citep{schaeffer2026}.

The remedies that work share one feature: each brings in information from
outside the self-consuming loop. Fresh real data \citep{alemohammad2024},
retained real data \citep{gerstgrasser2024}, verifiers of correctness
\citep{feng2025,yi2026}, correction toward ground truth
\citep{gillman2024}, and human curation \citep{ferbach2024} all import
something the model does not already have. And the theory that supports
them shares two assumptions. The outside information is treated as an
exogenous archive, there to be drawn on in whatever quantity is needed.
And the target distribution is fixed: what the model should know does not
change while it is being copied.

\begin{revised}
Neither assumption holds for the systems that motivate the problem. The
fresh text on which future models are trained is written by people, and
people increasingly consult the models instead of finding things out for
themselves; the supply of fresh data responds to the models it feeds. And
the world that text describes keeps changing. Cultural evolution has
studied both features for decades, and it separates two ways in which
copying loses knowledge. In a fixed world, each copy is imperfect, and a
population that learns only by copying loses its rarest skills unless
enough people keep supplying good models \citep{henrich2004}. In a world
that changes, copies are also outdated: Rogers' paradox
\citep{rogers1988} is that copying spreads because it is cheaper than
learning, and spreads until the population is no better off than one that
never copied, because someone must keep paying to learn what is true now.
Both are answers to the question this paper asks of model collapse: who
supplies the knowledge that a lineage of models copies, and what does it
cost them?

\paragraph{Terms.} A \emph{lineage} is a sequence of models in which
generation $t+1$ is trained on a corpus that mixes a fraction $f$ of
\emph{fresh} data, written by people about the world, with a fraction
$1-f$ of \emph{synthetic} data, sampled from generation $t$. In a
\emph{fixed world} the source of fresh data stays the same; in a
\emph{changing world} it changes between generations, either because the
text people write moves on (Section~\ref{sec:textdrift}) or because facts
change their value (Section~\ref{sec:factworld}).

\paragraph{Contributions.}
(i)~We write lossy transmission and Rogers' staleness as one chain whose
parameters (what fresh data supply, how faithfully copies transmit, and
how fast the world changes) can be measured separately
(Section~\ref{sec:theory}). (ii)~In a fixed world, a zero-parameter
neutral-drift baseline predicts the rare-word coverage of word-trigram
lineages; the loss is Henrich's, not Rogers'. A pretrained transformer
deviates from it in a way that points to its pretrained weights as a
reserve outside the chain (Section~\ref{sec:static}). (iii)~In a world of
facts that change at a set rate, we show Rogers' paradox at the level of
the model: accuracy falls and outdated answers rise with the rate of
change, the chain calibrated on the static world predicts the changing one,
and inheriting weights helps in a static world and carries stale facts in
a changing one (Section~\ref{sec:changing}). (iv)~Closing the loop with the
people who write the data, we derive the share of people who must keep
learning, in a fixed and in a changing world, extend it to heterogeneous
costs, aggregating models, and users who check answers, and show when
checking is stable (Section~\ref{sec:ecosystem}). (v)~We place
synthetic-data filters among social learning strategies, and show that
filters selecting plausible text act as conformist copying
(Section~\ref{sec:filters}).
\end{revised}

\section{Background: two ways copying loses knowledge}
\label{sec:background}

\begin{revised}
\paragraph{Lossy transmission.} \citet{henrich2004} modelled a population
that acquires a skill by copying the most skilled member of the previous
generation. Each copy falls short on average, and only occasionally does a
copier exceed its model; whether the skill is maintained or lost depends
on how many people try, because more attempts make an improving copy more
likely. When a population shrinks, complex skills are lost even though
nothing in the environment has changed. In population genetics the same
structure appears as neutral drift: a finite sample of the previous
generation loses rare variants by chance, and migration from outside
restores them \citep{wright1931,ewens2004}. A lineage of models that
samples its own training data is such a population, and fresh data are its
migrants.
\end{revised}

\paragraph{Rogers' paradox.} In Rogers' model \citep{rogers1988}, the best
behaviour in an environment changes with probability $u$ per generation.
An individual learner pays a cost $c$ and acquires the currently correct
behaviour; a social learner copies a random member of the previous
generation at no cost, and is right only if its model was right and the
environment has not changed since. Social learning spreads while rare,
because its copies come mostly from learners, and stops spreading when
social and individual learners do equally well. At that point, the
population's mean payoff equals that of a population of individual
learners alone: copying has been adopted without making anyone better off.
\rev{The two mechanisms differ in what goes wrong with a copy: under lossy
transmission it is a worse version of something still true, under Rogers'
it is a faithful version of something no longer true.}

\paragraph{Strategies.} Three refinements matter here. A \emph{critical}
social learner copies, checks the copy against the world, and learns for
itself if the copy fails; such learners raise the population's payoff
above the baseline, because the check imports information
\citep{enquist2007}. A \emph{conformist} learner copies what is common
rather than a random model; conformity is cheap, but it erodes variation,
because rare variants are copied less often than they occur
\citep{boyd1985}. And a \emph{payoff-biased} learner copies whoever does
best, which concentrates the population on the current winner
\citep{boyd1985,laland2004}. The social learning strategies tournament
found that successful strategies copy most of the time but copy well
\citep{rendell2010}. Training a model on another model's output is social
learning in the sense of \citet{heyes2012}: the learner reconstructs
competence from another's behaviour through its own learning machinery.

\section{Related work}
\label{sec:related}

\begin{revised}
\paragraph{Social dynamics of AI and collapse.} \citet{collins2026}
simulate Rogers' paradox in a population that can also learn from an AI
system, which itself updates by tracking the population's mean. They find
the paradox intact and show that it is broken when people can override the
AI by relearning after a failed copy, a form of critical social learning;
they also study how often the AI should update and how reliance can erode
people's own ability to learn. What they explicitly set aside is the part
that defines model collapse, models trained on the output of other models,
which we take as central: our models have a second copying channel, the
share of model-written text, and our equilibria are in closed form (the
critical asker of Section~\ref{sec:ecosystem} is their override strategy).
\citet{baumann2026} model people choosing between effort and a generative
model whose benefit declines with use, and show that welfare can fall below
its pre-AI level; their model does not use social-learning theory or the
model-to-model loop. \citet{guo2026ce} and \citet{ren2024il} read collapse
through iterated learning, which explains which structures degrade and in
what order. \citet{peterson2024} argues that reliance on AI narrows public
knowledge, and \citet{brinkmann2023} set out the programme of studying
machines as participants in cultural evolution. None of these separates
lossy transmission from staleness, or tests a chain's quantitative
predictions on trained language models.

\paragraph{Continual learning and distillation.} A lineage that inherits
its weights is a continual learner on a stream whose later batches are
partly its own output. Continual learning studies how a network keeps old
knowledge while learning new data \citep{french1999,kirkpatrick2017}, and
continual knowledge learning asks the same of the facts a language model
holds when the world updates them \citep{jang2022}; temporal
generalization studies how models trained on the past degrade on the
future \citep{lazaridou2021,dhingra2022}. Our inherit lineages show both
sides of this trade-off in one system: the same weights that retain facts
across generations also retain facts that have since changed
(Section~\ref{sec:reserve}). Training a model on a model's output is also
distillation \citep{hinton2015}; repeated self-distillation can improve a
student \citep{furlanello2018} but progressively regularizes it toward a
narrower function \citep{mobahi2020}, which is the narrowing that collapse
studies observe over generations of sampling.

\paragraph{Knowledge in language models.} Language models learn facts in
proportion to how often they are seen and struggle with long-tail facts
\citep{kandpal2023}; synthetic worlds of fictional individuals have been
used to study how such facts are stored and extracted
\citep{allenzhu2023}. Our fact world follows that design, adds change over
generations, and measures not only whether a fact is known but whether an
outdated value is given in its place.
\end{revised}

\section{Theory: one chain, two costs}
\label{sec:theory}

\begin{revised}
Consider one unit of knowledge, such as a word type in a corpus or a fact
in a model, and a lineage with fresh fraction $f$. Generation $t+1$ holds
the unit if the fresh part supplies it, with probability $\phi$, or else if
the synthetic part transmits it, with probability $\sigma$, and it is
still true, with probability $1-u$. The share $q_t$ of such units that the
lineage holds then obeys
\begin{equation}
  q_{t+1} = \phi + (1-\phi)\,\sigma\,(1-u)\,q_t ,
  \qquad
  q^* = \frac{\phi}{1-(1-\phi)\,\sigma\,(1-u)} ,
  \label{eq:chain}
\end{equation}
and the share $z_t$ of units for which the lineage holds a value that was
true earlier but is false now obeys, to first order in $u$,
\begin{equation}
  z_{t+1} = (1-\phi)\,\sigma\,(u\,q_t + z_t) .
  \label{eq:stale}
\end{equation}
The three parameters have separate meanings. $\phi$ is what fresh data
alone supply; it depends on $f$ and is concave in it, because a unit of
frequency $\pi$ is absent from $n$ fresh tokens with probability
$(1-\pi)^n$. $\sigma$ is the fidelity of the copying channel. $u$ is the
rate at which the world changes. Eq.~\ref{eq:chain} contains both
mechanisms of Section~\ref{sec:background}:

\emph{Lossy transmission} ($u=0$, $\sigma<1$). Knowledge is lost although
nothing changes, $q^*=\phi/(1-(1-\phi)\sigma)<1$, and only fresh supply
($\phi>0$) prevents total loss. For a corpus whose synthetic part is
sampled from the previous corpus's frequencies, $\sigma$ is fixed by those
frequencies: the lineage is a Wright--Fisher population with fresh data as
migration, and its coverage can be predicted with no free parameter.

\emph{Rogers' staleness} ($u>0$). Even perfect copies ($\sigma=1$) lose
value, $q^*=\phi/(1-(1-\phi)(1-u))$, and the units lost are not missing
but wrong: Eq.~\ref{eq:stale} predicts outdated answers that grow with
$u$ and with the share of copied knowledge.

The first version of this paper used $q_{t+1}=(1-\rho)+\rho(1-u)q_t$ with
$\rho=1-f$, one rate $u$ fitted to fully synthetic training, and
$\phi=1-\rho$. As a reviewer pointed out, fresh data alone cover far less
than $1-\rho$ of the rare words (measured below), and in a fixed world $u$
is sampling loss, not change; the good fit of that form at the final
generation came from errors that cancel. Eq.~\ref{eq:chain} separates the
terms, and the experiments test it in three ways that need no fitting to
the condition predicted:

\begin{itemize}
\item[P1] In a fixed world, the coverage of a lineage's corpus follows
neutral drift with fresh data as migration (Section~\ref{sec:static}).
\item[P2] In a changing world, $\phi$ and $\sigma$ measured in the static
world, together with the set rate $u$, predict accuracy and outdated
answers (Section~\ref{sec:factworld}).
\item[P3] Knowledge kept outside the chain (pretrained weights restored at
every generation) is neither lost by copying nor outdated by it; knowledge
kept in inherited weights is copied with high fidelity, which helps when
$u=0$ and carries outdated values when $u>0$ (Section~\ref{sec:reserve}).
\end{itemize}
In both regimes the plateau requires $\phi>0$: someone must keep supplying
fresh knowledge. Section~\ref{sec:ecosystem} asks who.
\end{revised}

\section{A fixed world: lossy transmission}
\label{sec:static}

\paragraph{Setup.} We train lineages of language models on two corpora
from Project Gutenberg, Darwin's \emph{On the Origin of Species} and
Melville's \emph{Moby-Dick}. Each corpus is split by sentence into a test
set (10\%), never used for training; a reference set (20\%), used only by
one filter (Section~\ref{sec:filters}); and a pool (70\%) from which the
first corpus and all fresh data are drawn. The main model is a word trigram
language model interpolating trigram, bigram, and unigram estimates, chosen
because its knowledge can be counted exactly; we repeat the key conditions
with a small character-level transformer and with a pretrained
transformer language model. Generation~0 is trained on
30\,000 tokens (words and punctuation marks) from the pool. Each later
generation is trained on a new corpus of the same size, built from the
previous model's samples according to the condition, for 12 generations
and 5 seeds. \rev{Fresh data are drawn anew at every generation: the pool
is shuffled and whole sentences are taken until the token budget is met,
so within a generation sentences are drawn without replacement, while
across generations draws are independent and may repeat sentences of
earlier corpora, including generation~0; seeds differ in the shuffles and
in sampling (Appendix~\ref{app:details}).}

\paragraph{Measures.} \emph{Rare-word coverage}: of the words outside the
pool's 1\,000 most frequent, the share present in the training corpus,
relative to generation~0. \rev{This is a property of the corpus, not of the
model; model-level measures follow in Section~\ref{sec:changing}.}
\emph{Perplexity} on the human test set, as a growth factor over
generation~0. \emph{Variety}: distinct three-word sequences per 1\,000
words of the model's own samples.

\begin{revised}
\paragraph{Prediction.} The neutral-drift baseline replaces the trained
model by its word frequencies: the synthetic part of each corpus is
$(1-f)N$ tokens drawn i.i.d.\ from the previous corpus's word counts, and
the fresh part is real pool sentences drawn as above. It has no free
parameter. If the trigram lineage loses rare words faster than this, its
model adds loss beyond sampling; if slower, something re-supplies them.
\end{revised}

\begin{figure}[t]
  \centering
  \includegraphics[width=\textwidth]{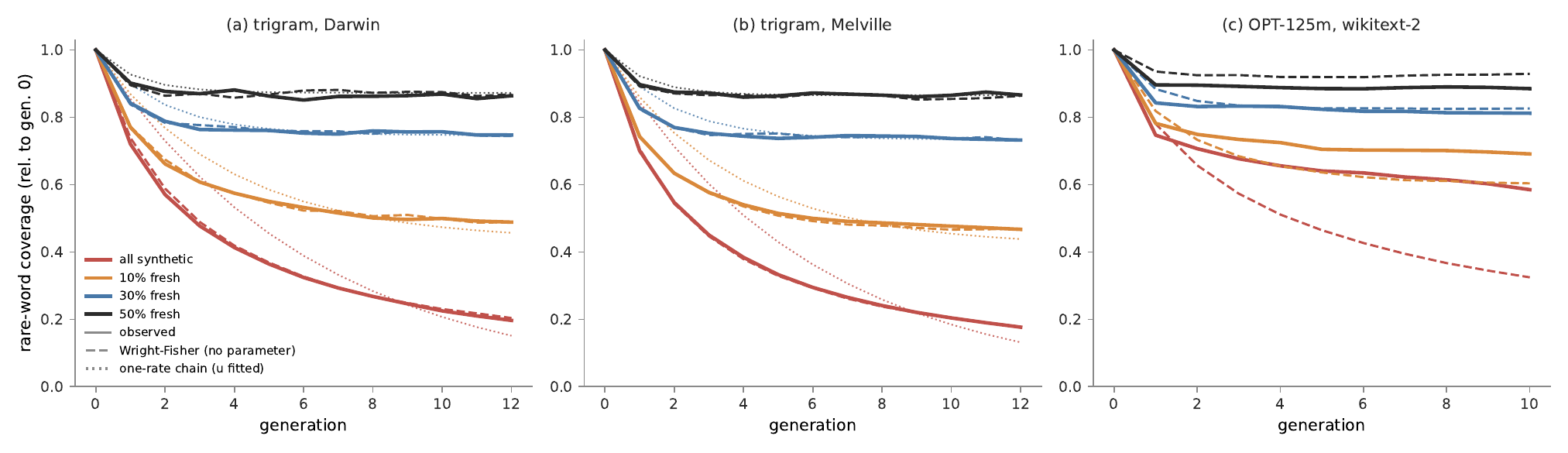}
  \caption{\rev{A fixed world is lossy transmission. Rare-word coverage of
  the training corpus relative to generation~0 (solid; mean of 5 seeds),
  the zero-parameter neutral-drift (Wright--Fisher) prediction (dashed),
  and the one-rate chain of the first version of this paper, its rate
  fitted to the all-synthetic condition (dotted). (a,~b)~Word trigram
  lineages. (c)~OPT-125m on wikitext-2: the pretrained model loses more
  than drift at the first copy and less afterwards.}}
  \label{fig:static}
\end{figure}

\begin{table}[t]
  \caption{Word trigram lineages, generation 12, mean of 5 seeds.
  Coverage: rare-word coverage relative to generation~0. \rev{Drift: the
  zero-parameter neutral-drift prediction. Fresh only: coverage of the fresh
  part alone.} Perplexity: growth factor on held-out human text.}
  \label{tab:static}
  \centering
  \small
  \begin{tabular}{lcccccccc}
  \toprule
  & \multicolumn{4}{c}{Darwin} & \multicolumn{4}{c}{Melville} \\
  \cmidrule(lr){2-5}\cmidrule(lr){6-9}
  condition & coverage & \rev{drift} & \rev{fresh only} & perplexity & coverage & \rev{drift} & \rev{fresh only} & perplexity \\
  \midrule
  all synthetic & 0.20 & \rev{0.20} & -- & 7.7 & 0.18 & \rev{0.18} & -- & 4.1 \\
  10\% fresh & 0.49 & \rev{0.49} & \rev{0.15} & 2.8 & 0.47 & \rev{0.47} & \rev{0.13} & 2.1 \\
  30\% fresh & 0.75 & \rev{0.75} & \rev{0.40} & 1.6 & 0.73 & \rev{0.73} & \rev{0.36} & 1.4 \\
  50\% fresh & 0.86 & \rev{0.86} & \rev{0.60} & 1.3 & 0.87 & \rev{0.87} & \rev{0.57} & 1.2 \\
  accumulate all past data & 1.00 & -- & -- & 1.3 & 1.00 & -- & -- & 1.4 \\
  \midrule
  filter: own likelihood & 0.24 & -- & -- & 7.9 & 0.21 & -- & -- & 4.6 \\
  filter: reference likelihood & 0.02 & -- & -- & 11.2 & 0.01 & -- & -- & 4.9 \\
  filter: n-gram match & 0.03 & -- & -- & 107.5 & 0.02 & -- & -- & 34.7 \\
  filter: random half & 0.20 & -- & -- & 7.7 & 0.18 & -- & -- & 4.2 \\
  filter: favour rare words & 0.56 & -- & -- & 12.3 & 0.44 & -- & -- & 7.3 \\
  \bottomrule
  \end{tabular}
\end{table}

\paragraph{Results.} Trained only on its own samples, the trigram model
loses four in five of its rare words in twelve generations, and its
perplexity on human text grows by a factor of 7.7 on Darwin and 4.1 on
Melville (Table~\ref{tab:static}); a sample from generation~0 reads ``i
cannot see that a genus become fitted for most justly celebrated elder
huber, but in all the large stock'', and one from generation~12 ``struggle
of the on that may inheritable detected i hybrids, kept.'' Any fresh
fraction halts the decline. \rev{Neutral drift predicts every trajectory,
with no parameter: the largest error at generation~12 is 0.007, and at any
generation 0.028, over all conditions and both corpora
(Fig.~\ref{fig:static}a,b). The one-rate chain of the first version errs
by up to 0.17 during the transient, and its accurate plateaus rest on two
errors that cancel: it assumes that 10\%, 30\%, and 50\% fresh data alone
supply 0.10, 0.30, and 0.50 of the rare words, whereas they supply 0.15,
0.40, and 0.60 on Darwin (0.13, 0.36, 0.57 on Melville;
Table~\ref{tab:static}), and it underestimates what copying keeps.} For a
trigram model, then, collapse in a fixed world is the loss of rare variants
by sampling, which fresh data replace as migration does: Henrich's lossy
transmission, not Rogers' staleness. Keeping every past corpus avoids
collapse, as \citet{gerstgrasser2024} found; in these terms the original
corpus is a source of migrants that never leaves.

\begin{figure}[t]
  \centering
  \includegraphics[width=0.85\textwidth]{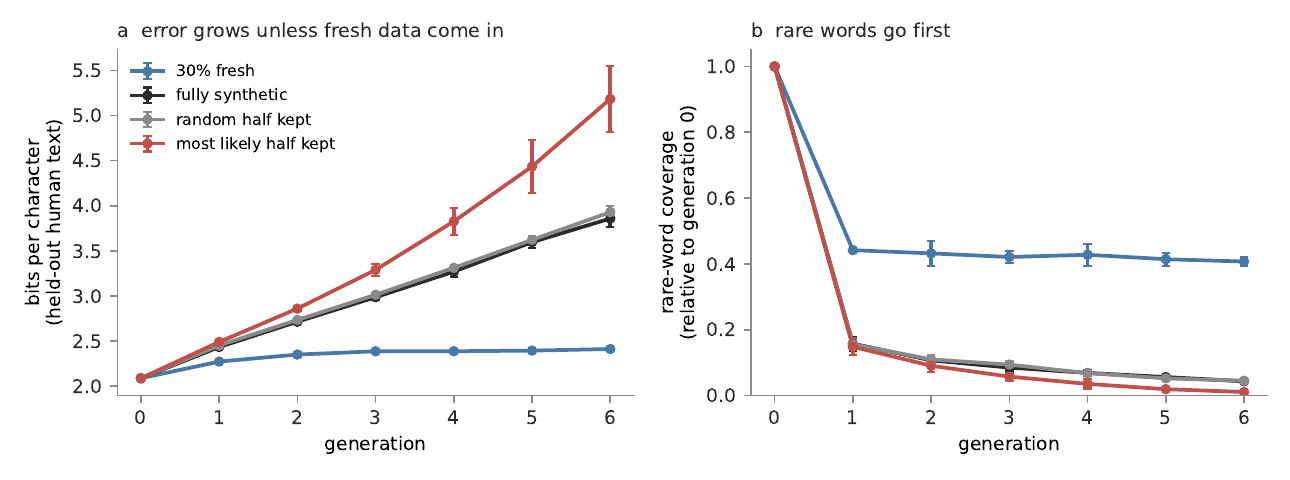}
  \caption{Recursive training of a small character-level transformer
  (Darwin; 6 generations; mean $\pm$ s.d.\ over 3 seeds). (a)~Bits per
  character on held-out human text. (b)~Rare-word coverage of the training
  corpus, relative to generation 0. Fresh data stop the decline; keeping
  the most likely half of the samples accelerates it relative to keeping a
  random half.}
  \label{fig:neural}
\end{figure}

\paragraph{A neural replication.} To check that the pattern is not a
property of count-based models, we repeated four conditions with a small
character-level transformer (2 layers, width 128, context 128 characters)
trained afresh each generation on 60\,000 characters of Darwin's text and
sampled at temperature 1 (6 generations, 3 seeds;
Fig.~\ref{fig:neural}). The qualitative picture is the same. Under fully
synthetic training, error on held-out human text rises at every generation
(1.77 more bits per character by generation 6), and rare-word coverage
falls to 4\% of its starting value; with 30\% fresh data, both settle
within two generations (coverage 0.41 of its starting value at generation
6, compared with 0.04; Welch $p=0.002$, Cohen's $d=16.5$). The loss is
front-loaded: a character model sampled at temperature 1 misspells most
rare words in its first generation, so its copying channel is far less
faithful than word sampling. Filters are discussed in
Section~\ref{sec:filters}. With three seeds per condition the rank-based
tests cannot reach $p<0.05$ (Mann--Whitney $p=0.08$--$0.10$ for every
comparison); we report them for completeness.

\begin{figure}[t]
  \centering
  \includegraphics[width=\textwidth]{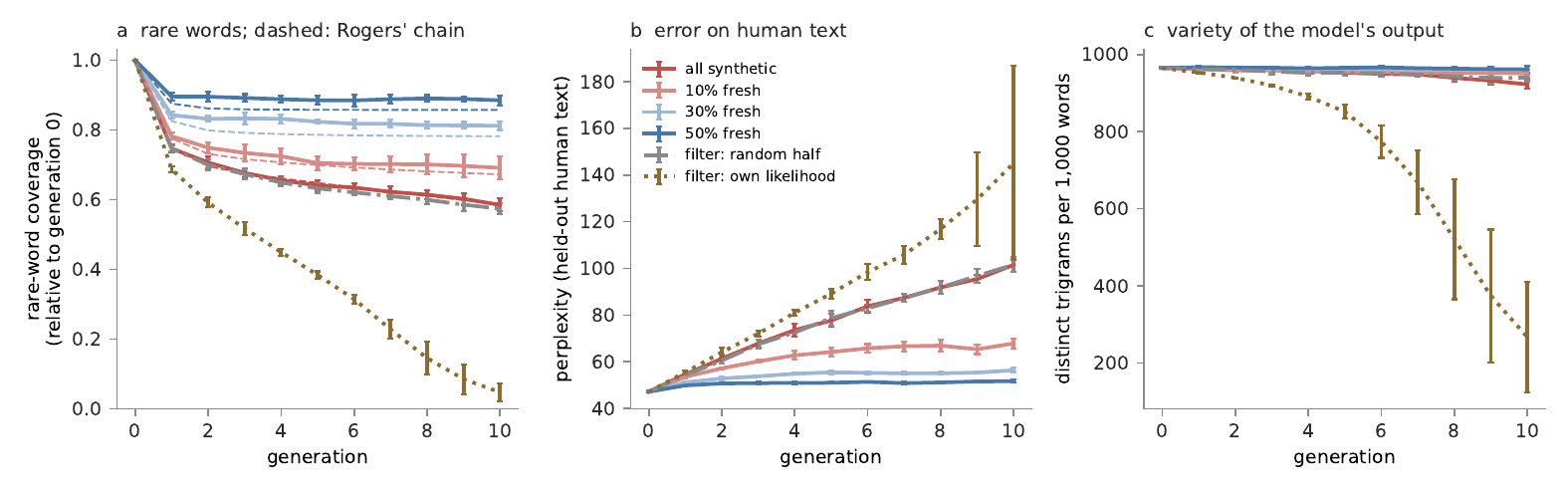}
  \caption{Recursive fine-tuning of a pretrained language model (OPT-125m
  on wikitext-2; 10 generations; mean $\pm$ s.d.\ over 5 seeds).
  (a)~Rare-word coverage of the training corpus, relative to generation 0,
  with the two-class chain of Appendix~\ref{app:chain} (dashed).
  (b)~Perplexity on held-out human text. (c)~Distinct word trigrams per
  1\,000 words of the model's own samples. Keeping the samples the model
  finds most likely collapses the lineage; keeping a random half does
  nothing.}
  \label{fig:opt}
\end{figure}

\paragraph{A pretrained language model.} We repeated the static
experiment in the setting of \citet{shumailov2024}: OPT-125m
\citep{zhang2022opt}, fine-tuned afresh from its pretrained weights at
every generation (2 epochs) on 4\,000 blocks of 64 tokens from wikitext-2
\citep{merity2017}, with each generation's synthetic blocks sampled from
the start token at temperature 1, so no human text enters through prompts
(10 generations, 5 seeds per condition; Fig.~\ref{fig:opt}). Trained only
on its own output, the lineage doubles its perplexity on human text (47 to
102) and keeps 0.59 of its rare words; with 10\%, 30\%, and 50\% fresh
blocks, perplexity levels off at 68, 56, and 52 and coverage at 0.69,
0.81, and 0.89 (30\% fresh against fully synthetic at generation 10:
coverage difference 0.23, Welch $p<10^{-6}$, Mann--Whitney $p=0.008$,
Cohen's $d=14.6$, bootstrap 95\% CI $[0.21, 0.24]$). \rev{Here neutral
drift does not suffice (Fig.~\ref{fig:static}c). With 30\% and 50\% fresh
data it predicts the plateau to within 0.01 and 0.04, but the lineage
departs from it in both directions: it loses more at the first copy (0.75
against 0.78 retained), and, trained only on its own output, it keeps far
more by generation 10 (0.59 against 0.33). Something outside the corpus
re-supplies rare words. The obvious candidate is the pretrained weights,
which every generation in this design starts from: in Henrich's terms a
source of innovation that does not pass through the copying chain. We test
this directly in Section~\ref{sec:reserve}.}

\begin{revised}
\paragraph{What is new here.} Three fixed-world findings confirm prior
work: fresh data stop the decline \citep{alemohammad2024,shumailov2024},
accumulating data prevents collapse \citep{gerstgrasser2024}, and
likelihood-based selection speeds the loss of diversity
\citep{alemohammad2024,ferbach2024}. What the cultural-evolution reading
adds is the identification of the mechanism with a zero-parameter
baseline, which turns ``how much fresh data is needed'' into a computation
on word frequencies for count-based models, and a way to see when a model
departs from it: the pretrained transformer departs in the direction of a
reserve outside the chain.
\end{revised}

\section{A changing world: Rogers' paradox}
\label{sec:changing}

\rev{Rogers' paradox requires a world that changes. We let it change in two
ways: the text people write moves on, and facts change their value.}

\subsection{Text that moves on}
\label{sec:textdrift}

\paragraph{Setup.} The world at generation $t$ is a window of 3\,000
consecutive sentences in the concatenated text of Darwin's and Melville's
books, which slides forward by a fixed number of sentences per generation,
wrapping around; new vocabulary therefore keeps arriving and old
vocabulary keeps leaving. Fresh data are drawn from the current window,
\rev{with replacement,} and every tenth sentence of the window is held
out. Each generation is trained on 30\,000 tokens, a fraction $f$ fresh and
the rest sampled from the previous model. Coverage is measured against the
current window's rare words, relative to a corpus of the same size drawn
entirely fresh from the current window, and averaged over generations 6 to
12, after the initial transient (3 seeds).

\begin{figure}[t]
  \centering
  \includegraphics[width=\textwidth]{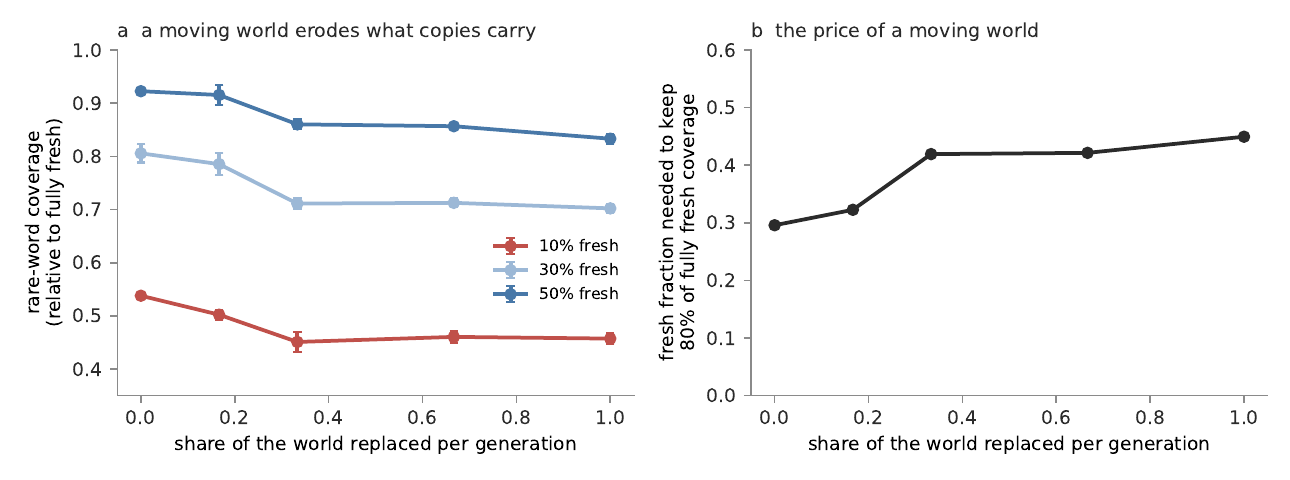}
  \caption{Recursive training in a world that keeps changing (word trigram
  models; the world is a window of 3\,000 sentences sliding through the
  concatenated corpora; mean of generations 6--12 and 3 seeds; bars: s.d.
  over seeds). (a)~Coverage of the current world's rare words, relative to
  a fully fresh corpus, against the share of the world replaced per
  generation. (b)~The fresh fraction needed to keep 80\% of fully fresh
  coverage rises from about 0.30 in a static world to about 0.45 when the
  world is replaced every generation.}
  \label{fig:drift}
\end{figure}

\paragraph{Results.} At every fresh fraction, coverage falls as the world
changes faster (Fig.~\ref{fig:drift}a): with 30\% fresh data, from 0.81 in
a static world to 0.70 when the window is replaced every generation; with
10\%, from 0.54 to 0.46. Equivalently, the fresh fraction needed to retain
80\% of what a fully fresh corpus would cover rises from about 0.30 to
about 0.45 (Fig.~\ref{fig:drift}b). \rev{The cause is measurable: the share
of the current window's rare words that were absent from the previous
window rises from 0 to 0.09, 0.21, 0.40, and 0.56 as the window slides by
500 to 3\,000 sentences, and the copied part of the corpus cannot contain
them. But this is not yet Rogers' mechanism. Neutral drift on the moving
window, again with no parameter, matches the trigram lineage's late
coverage in all 15 combinations of speed and fresh fraction to within
0.009: new words are missing, not contradicted. Text that moves on adds
vocabulary; it rarely makes what was copied false.} The same holds for
OPT-125m: on wikitext-2 windows of 6\,000 blocks sliding by up to the whole
window per generation, coverage of the current window's rare words is the
same whether the world stands still or is replaced (with 30\% fresh data,
0.698 against 0.700, Welch $p=0.66$; 5 seeds), and perplexity on the
current world rises only slightly (46.0 against 45.0).

\subsection{Facts that change}
\label{sec:factworld}

\begin{revised}
\paragraph{Setup.} To test Rogers' mechanism we need copies that become
false. We built a world of 2\,000 fictional people (random pronounceable
names, so that the pretrained model has no prior about them), each living
in one of 50 cities whose names are single tokens. Human text at
generation $t$ states the current truth in 16 paraphrase templates (``X
lives in Y.'', ``Y is home to X.'', ``X took the train back to Y.'',
\ldots), with people mentioned at Zipf frequencies (exponent~1), so that
the 200 most mentioned (\emph{head}) appear hundreds of times per corpus,
the next 800 (\emph{mid}) a few times, and the last 1\,000 (\emph{tail})
about once. Between generations, each fact changes to another city with
probability $u$, which we set. Lineages are as before: OPT-125m, 4\,000
blocks of 64 tokens per generation, a fraction $f$ fresh (rendered from the
current truth) and the rest sampled from the previous model, 10
generations (Appendix~\ref{app:details}). We use two designs. In
\emph{restart} lineages every generation fine-tunes the pretrained
weights, as above; in \emph{inherit} lineages every generation continues
from the previous generation's weights. The grid is $u\in\{0, 0.05, 0.1,
0.2, 0.4\}$, $f\in\{0.1, 0.3, 0.5\}$, both designs, and a fully fresh
lineage for each $u$, with 3 seeds (105 lineages).

\paragraph{Measures.} We ask the model ``X lives in'' and three other
forward templates and average its next-token distribution over the 50
cities. \emph{Accuracy}: the most probable city is the current one.
\emph{Stale answers}: the most probable city was this person's city at an
earlier generation and is not now. Both are properties of the model,
measured on all 2\,000 people, and are averaged over generations 6 to 10.

\begin{figure}[t]\revstart
  \centering
  \includegraphics[width=\textwidth]{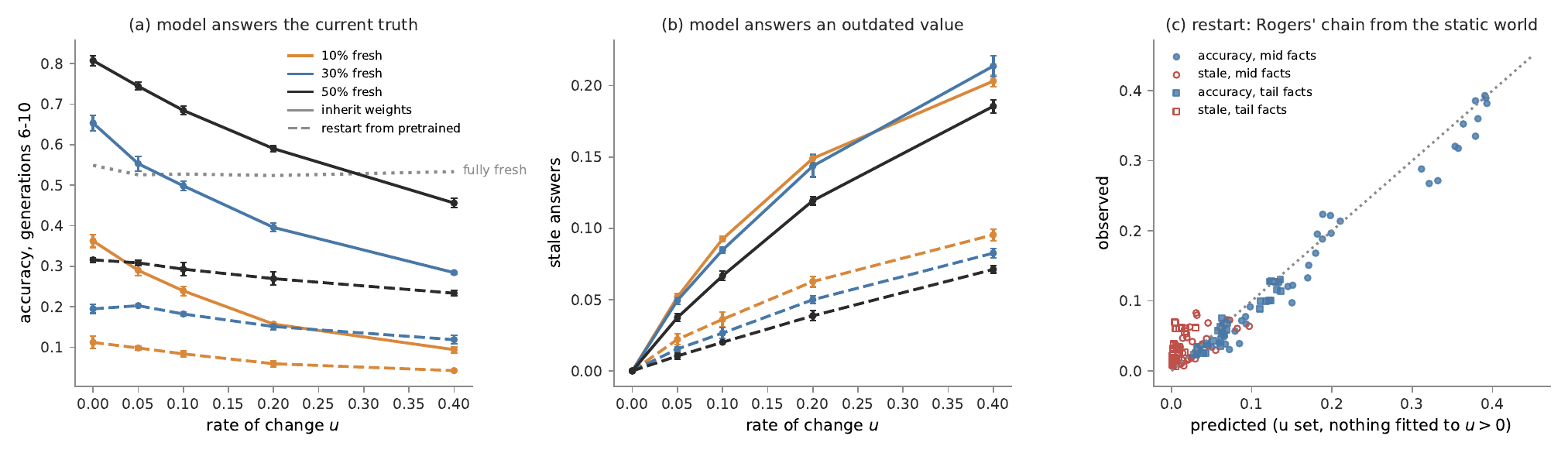}
  \caption{Rogers' paradox at the level of the model (OPT-125m in a world
  of 2\,000 fictional facts; generations 6--10; mean $\pm$ s.d.\ over 3
  seeds). (a)~Accuracy against the rate $u$ at which facts change; solid:
  inherit weights, dashed: restart from pretrained weights; dotted: fully
  fresh lineage. (b)~Answers that name a city that was true earlier and is
  false now. (c)~Restart lineages: Rogers' chain calibrated on the static
  world ($u=0$) predicts accuracy and stale answers for $u>0$ with $u$ set
  (each point: one fresh fraction, frequency class, seed, and $u$; mid and
  tail facts).}
  \label{fig:fact}
\end{figure}

\begin{table}[t]\revstart
  \caption{Fact world: accuracy on all 2\,000 facts, generations 6--10, mean
  $\pm$ s.d.\ over 3 seeds; in brackets, the share of stale answers.}
  \label{tab:fact}
  \centering
  \small
  \begin{tabular}{lccccccc}
  \toprule
  & \multicolumn{2}{c}{10\% fresh} & \multicolumn{2}{c}{30\% fresh} & \multicolumn{2}{c}{50\% fresh} & fully \\
  \cmidrule(lr){2-3}\cmidrule(lr){4-5}\cmidrule(lr){6-7}
  $u$ & restart & inherit & restart & inherit & restart & inherit & fresh \\
  \midrule
  0    & .11 [.00] & .36 [.00] & .19 [.00] & .65 [.00] & .32 [.00] & .81 [.00] & .55 \\
  0.05 & .10 [.02] & .29 [.05] & .20 [.02] & .55 [.05] & .31 [.01] & .74 [.04] & .53 \\
  0.1  & .08 [.04] & .24 [.09] & .18 [.03] & .50 [.08] & .29 [.02] & .68 [.07] & .53 \\
  0.2  & .06 [.06] & .16 [.15] & .15 [.05] & .40 [.14] & .27 [.04] & .59 [.12] & .52 \\
  0.4  & .04 [.10] & .09 [.20] & .12 [.08] & .28 [.21] & .23 [.07] & .46 [.19] & .53 \\
  \bottomrule
  \end{tabular}
\end{table}

\paragraph{Results.} In a static world ($u=0$) a restart lineage loses most
of what a fully fresh lineage knows (with 30\% fresh data, 0.19 against
0.55; Table~\ref{tab:fact}), and it gives no stale answers: this is lossy
transmission, now at the level of the model. As facts start to change,
accuracy falls and stale answers rise in every combination of design and
fresh fraction (Fig.~\ref{fig:fact}a,b), while the fully fresh lineage
stays near 0.53. The answers lost are not blanks: with 30\% fresh data and
$u=0.4$, a restart lineage names an outdated city for 8\% of people and an
inherit lineage for 21\% (Welch $p=2\times10^{-4}$), against at most 3\% for
the fully fresh lineage. This is Rogers' mechanism: copies are faithful
reports of a world that no longer exists, and they cost the lineage in
proportion to how fast the world moves.

\paragraph{Prediction.} For restart lineages we calibrated $\phi$ and
$\sigma$ of Eq.~\ref{eq:chain} on the static lineage of each fresh
fraction, frequency class, and seed, and then predicted every $u>0$ with
$u$ set and nothing fitted (Eq.~\ref{eq:chain} for accuracy,
Eq.~\ref{eq:stale} for stale answers; 72 cells for mid and tail facts).
For mid-frequency facts the mean absolute error is 0.023 for accuracy
(largest 0.060) and 0.015 for stale answers (largest 0.052); for tail
facts, 0.009 (largest 0.026) and 0.025 (largest 0.066)
(Fig.~\ref{fig:fact}c). For head facts the chain cannot be calibrated:
their static accuracy barely declines over ten generations, so $\phi$ and
$\sigma$ are not separately identifiable, and predictions from such fits
are off by up to 0.7. The chain describes inherit lineages poorly
(Section~\ref{sec:reserve}).
\end{revised}

\subsection{Where knowledge is kept: pretrained, inherited, copied}
\label{sec:reserve}

\begin{revised}
The reviewers of the first version asked whether the insensitivity of the
OPT lineage to a moving world was due to its pretrained weights, and
suggested testing text written after pretraining and lineages that
inherit their weights. We did both.

\paragraph{Text after the pretraining cut-off.} We collected 20\,700
abstracts of arXiv cs.CL papers, 450 per month from January 2023 to
October 2026, all written after OPT's pretraining data
\citep{zhang2022opt}, in submission order (77\,904 blocks of 64 tokens).
The world is a window of 6\,000 blocks that either stands still or is
replaced at every generation, so that by generation 10 the lineage is
trained on text from more than three years later. Lineages restart or
inherit as in the fact world, with 30\% fresh data (10 generations, 4
seeds). Because coverage is a corpus measure, we measure the model
directly: the negative log-likelihood (NLL) of the current window's rare
words (outside its 1\,000 most frequent) where they occur in held-out human
text, with the 1\,000 most frequent words as a control, each reported as
the excess over a fully fresh lineage on the same window and seed.

\begin{figure}[t]\revstart
  \centering
  \includegraphics[width=0.8\textwidth]{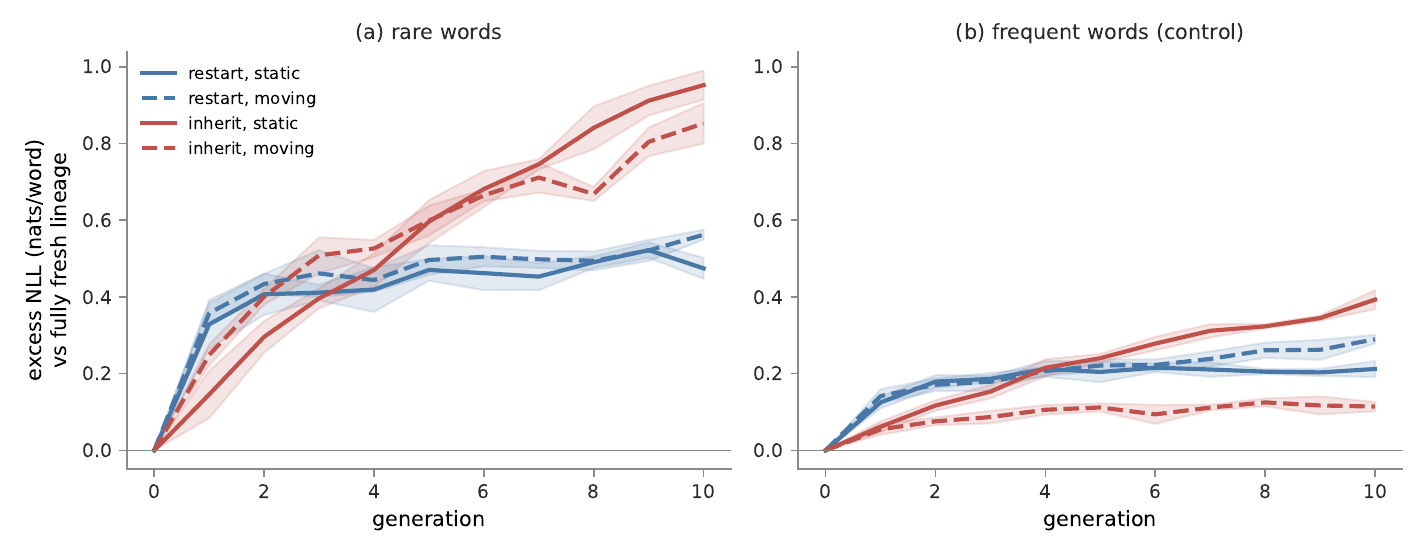}
  \caption{OPT-125m on arXiv cs.CL abstracts written after its pretraining
  (2023--2026; 30\% fresh; mean $\pm$ s.d.\ over 4 seeds). Excess negative
  log-likelihood of held-out human text over a fully fresh lineage on the
  same window: (a)~rare words, (b)~frequent words. Restart lineages level
  off; inherit lineages keep losing. Replacing the world every generation
  (dashed) adds little.}
  \label{fig:reserve}
\end{figure}

\paragraph{Results.} The model, not only its corpus, loses rare words
first: in late generations a restart lineage's excess NLL is 0.48 nats per
rare word against 0.21 per frequent word (difference 0.27, Welch
$p=4\times10^{-6}$, Cohen's $d=20$; Fig.~\ref{fig:reserve}). Inheriting
the weights makes it worse, not better: 0.83 nats (difference to restart
0.35, Welch $p=2\times10^{-6}$, bootstrap 95\% CI $[0.32, 0.37]$), still
rising by 0.07 nats per generation in every seed, while restart lineages
have levelled off (0.007 per generation). So the pretrained weights are a
reserve against lossy transmission, as the wikitext lineage suggested: a
lineage that starts from them every generation keeps a floor, and one that
carries its own weights forward erodes it, the forgetting that continual
learning studies \citep{french1999,kirkpatrick2017}. But replacing the
world at every generation, with text the pretrained model has never seen,
still adds little: 0.04 nats for restart lineages ($p=0.03$) and $-0.09$
for inherit lineages. The insensitivity to a moving world therefore does
not come from the reserve being up to date; it comes from text drift
itself, which, as for trigrams, adds what is new without making what was
copied false.

\paragraph{Inheritance in the fact world.} Where facts do change, inherited
weights behave as individual memory does in Rogers' theory. In a static
world they are an asset: inherit lineages accumulate facts across
generations and, with 30\% or 50\% fresh data, exceed even the fully fresh
lineage (0.65 and 0.81 against 0.55; Table~\ref{tab:fact}), because each
generation adds what it reads to what it already holds. As the world
changes, the asset turns into stale knowledge: inherit lineages lose
accuracy four times as fast as restart lineages (slope $-0.87$ against
$-0.21$ per unit of $u$ with 30\% fresh data, consistent across seeds),
give two to four times as many stale answers, and their advantage over
restart lineages shrinks from 0.46 at $u=0$ to 0.17 at $u=0.4$. This is
also why Eq.~\ref{eq:chain} fits them poorly: inherited weights are not a
text-copying channel with a fixed fidelity, and fresh data overwrite old
values in place. The three places where a lineage can keep knowledge
therefore behave as Section~\ref{sec:theory} predicts (P3): pretrained
weights restored at every generation hold knowledge that copying neither
loses nor outdates, but they hold nothing about a new world; inherited
weights keep both what is still true and what no longer is; and text
copied between generations loses knowledge in a fixed world and outdates it
in a changing one.
\end{revised}

\section{Closing the loop: who supplies fresh data?}
\label{sec:ecosystem}

The analyses above, and the literature, treat fresh data as a resource
outside the system. It is not. Fresh text is written by people, and
people who can ask a model instead of finding something out will often do
so. We now couple the two loops. \rev{The results of this section hold
within the model stated here; they are derived and checked by simulation
of the same model, and neither the equilibria nor the benefit of checking
have been tested with trained models or real users.}

\paragraph{Model.} Knowledge consists of many independent facts, each of
which changes to a new value with probability $u$ per generation, where a
generation is one model release. A fraction $h$ of people learn each fact
they need for themselves, at cost $c$, and are always right; the rest ask
the current model, at negligible cost, and are right with probability $a_t$,
the model's accuracy at the time of use. What people write becomes the
training corpus for the next model, together with a share $s$ of text the
model writes directly. Learners write what they found; askers write what
they were told. \rev{Text copied from the model keeps a fact with
probability $\sigma$ (lossy transmission; $\sigma=1$ in the first version
of this paper).} People adjust their strategy toward whichever pays more.
\rev{We assume \emph{probability matching}: the model states a fact
correctly with probability equal to the share of its corpus that states it
correctly. We relax this, and the assumption of equal costs, below.}

\begin{proposition}
\label{prop:eco}
Let $f = (1-s)\,h$ be the fresh fraction of the corpus. The model's
accuracy at use obeys \rev{$a_{t+1} = (1-u)\,[\,f + (1-f)\,\sigma\,a_t\,]$}.
If $0<h^*<1$ at equilibrium, then
\begin{equation}
  a^* = 1-c, \qquad
  \rev{f^* = \frac{(1-c)\,\bigl(1-(1-u)\,\sigma\bigr)}{(1-u)\,\bigl(1-(1-c)\,\sigma\bigr)}}, \qquad
  h^* = \frac{f^*}{1-s}, \qquad
  \bar W^* = 1-c ,
  \label{eq:eco}
\end{equation}
where $\bar W^*$ is mean payoff. An interior equilibrium exists only while
$s < 1 - f^*$. \rev{For $\sigma=1$, $f^*=u(1-c)/(c(1-u))$.}
\end{proposition}

\begin{proof}
Askers earn $a$ and learners $1-c$; an interior equilibrium of payoff-driven
strategy change requires $a^* = 1-c$. At the fixed point of the accuracy
recursion, \rev{$a^* = (1-u)f/(1-(1-u)(1-f)\sigma)$}; setting it to $1-c$
and solving for $f$ gives $f^*$. Mean payoff is $h^*(1-c)+(1-h^*)a^* = 1-c$.
\end{proof}

\begin{revised}
\begin{corollary}
\label{cor:fixed}
In a fixed world ($u=0$) with lossy copying ($\sigma<1$), the fresh
fraction at equilibrium is $f^*=(1-c)(1-\sigma)/(1-(1-c)\sigma)>0$.
Someone must keep paying to learn even when nothing changes.
\end{corollary}

The corollary answers the question of the title for both mechanisms: in a
fixed world people pay to replace what copying loses, in a changing world
also to replace what copying outdates, and the two costs add
(Fig.~\ref{fig:eco2}a). In both, the result $a^*=1-c$ is Rogers'
indifference condition: whatever the model's architecture or data volume,
its accuracy at use is set by the cost of the alternative it displaces.
\end{revised}

Proposition~\ref{prop:eco} has three consequences for machine learning,
\rev{within this model}. \emph{First}, at equilibrium the model is exactly
as reliable as it is costly for people to find out for themselves.
\emph{Second}, the population is no better off than it would be without the
model: this is Rogers' paradox for the human-AI ecosystem \rev{(with equal
costs; see below)}. \emph{Third}, text written by models is paid for in
human effort. The share of people who must keep learning for themselves
rises as $1/(1-s)$ with the share of model-written text, and beyond
$s = 1-f^*$ the model is no longer worth asking at all.

\begin{figure}[t]
  \centering
  \includegraphics[width=\textwidth]{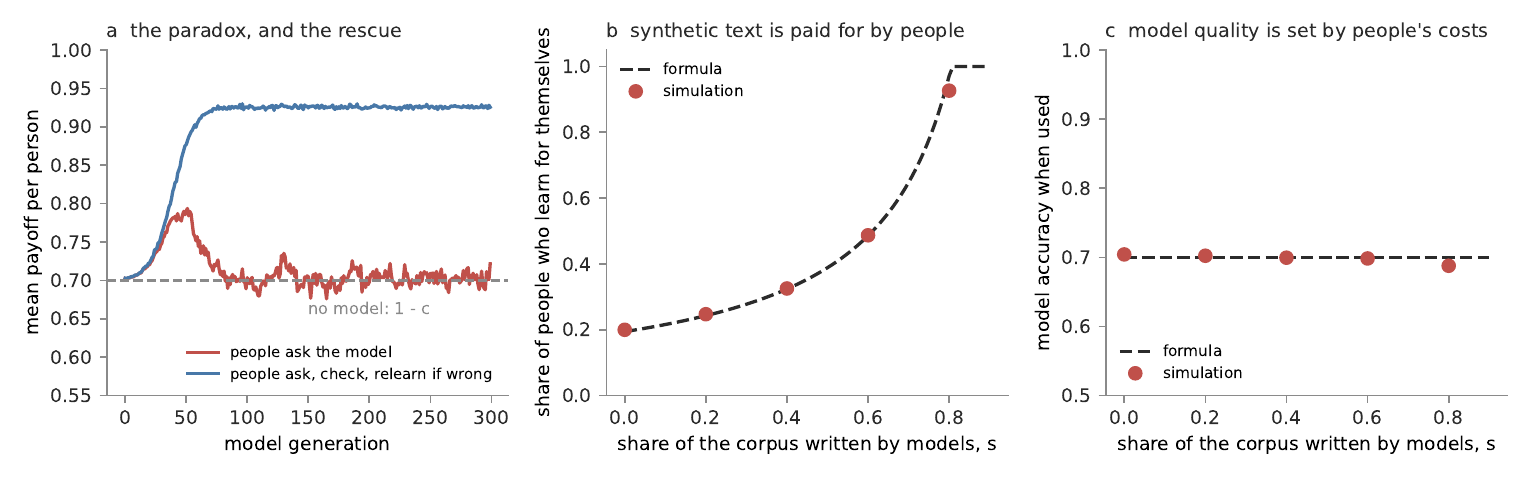}
  \caption{The human-AI knowledge ecosystem, simulated (4\,000 facts, 4\,000
  people, change rate $u=0.05$, cost of finding out $c=0.3$, checking cost
  $0.05$). (a)~Mean payoff after a model is introduced: when people simply
  ask, payoff rises and then returns to the no-model baseline $1-c$; when
  they check the answer and relearn if it is wrong, it stays well above.
  (b)~Equilibrium share of people who learn for themselves, against the
  share $s$ of model-written text: simulation and Eq.~\ref{eq:eco}
  \rev{with the coverage correction $\kappa$ derived in
  Appendix~\ref{app:proofs}}. (c)~The model's accuracy when used stays at
  $1-c$ whatever $s$.}
  \label{fig:eco}
\end{figure}

\paragraph{Simulation.} An agent-based simulation of the same model, with
strategy change by payoff-proportional imitation and a small mutation rate,
confirms the equilibrium (Fig.~\ref{fig:eco}). After a model is
introduced, mean payoff first rises, while most people still learn and the
model is accurate, and then falls back to $1-c$ as reliance spreads and the
model's corpus fills with copies (Fig.~\ref{fig:eco}a). The share of
learners matches Eq.~\ref{eq:eco} across synthetic shares from 0 to 0.6,
and the model's accuracy at use stays at $1-c$ throughout
(Fig.~\ref{fig:eco}b,c). \rev{In the simulation each person writes about
one random fact, so a fact receives any human-written text in a generation
only with probability $\kappa=1-(1-1/K)^N\approx 1-e^{-N/K}$ ($0.63$ for
$N=K$), and the fresh fraction is $f=(1-s)\kappa h$; hence $h^* =
f^*/((1-s)\kappa)$ (Appendix~\ref{app:proofs}).}

\begin{revised}
\paragraph{Heterogeneous costs.} If people differ in what learning costs
them, those with cost below $1-a$ learn and the rest ask, and the
indifference condition applies to the marginal learner: $a^*=1-c_m$, where
$c_m$ is the cost of the person indifferent between the two. People who
learn are no better or worse off than without the model; people for whom
learning is costlier than $c_m$ gain $a^*-(1-c)$. With costs uniform on
$[0, 0.6]$ (mean 0.3) and $s=0.3$, the model's accuracy settles at 0.81,
the marginal cost at 0.19, and mean payoff at 0.84 against 0.70 without the
model (Fig.~\ref{fig:eco2}b). So the strict paradox, no one better off,
depends on equal costs; what survives heterogeneity is that the model's
reliability is pinned by the marginal learner, that the gains go to those
for whom learning is most costly, and that more model-written text raises
the marginal cost and lowers the model's accuracy (0.84, 0.81, 0.76 at
$s=0$, 0.3, 0.6).

\paragraph{Aggregation.} A model need not match probabilities. If it
states the majority of what its corpus says about a fact, a fact that
changes is answered wrongly until fresh statements of the new value form a
majority, so the model is more stubborn. We simulated this per fact (each
corpus holds nine statements per fact; Appendix~\ref{app:proofs}). The
indifference result is unchanged, $a^*=0.70=1-c$ and $\bar W^*=1-c$, but
the fresh fraction needed rises from 0.12 under probability matching (the
formula gives 0.123) to 0.32 (Fig.~\ref{fig:eco2}c). Aggregation does not
break the paradox; it raises its price.

\paragraph{Checking, and whether it is stable.} If people who ask the
model also check its answer against the world, at a cost $c_v$ below that
of finding out, and learn for themselves when it is wrong (critical social
learning; \citealp{enquist2007}), their payoff is $1-c_v-(1-a)c$, and what
they write is correct.

\begin{proposition}
\label{prop:check}
With learners, askers, and checkers, if $c_v < c(1-c)$ and
$a^{**} = 1 - c_v/(1-c)$ can be sustained when everyone checks, the
population settles on askers and checkers only, at model accuracy
$a^{**}$ and mean payoff $\bar W^{**}=a^{**}>1-c$, and this equilibrium is
stable; if everyone checking cannot sustain $a^{**}$, everyone checks. If
$c_v\ge c(1-c)$, checkers do not outperform learners at $a=1-c$, and the
equilibrium of Proposition~\ref{prop:eco} is restored.
\end{proposition}

The proof is in Appendix~\ref{app:proofs}. With $c=0.3$ and $c_v=0.05$ the
proposition gives $\bar W^{**}=0.929$, which matches the agent-based
simulation (0.93, Fig.~\ref{fig:eco}a) and replicator dynamics from four
different starting mixtures for every $c_v$ we tried
(Fig.~\ref{fig:eco2}d). Cheap checking is thus not only a way out of the
paradox but a stable one, and its benefit is set by the ratio of checking
to learning costs. The implication for the design of AI systems is direct:
tools that make it cheap for users to verify an answer against a source
are the mechanism that keeps the ecosystem's knowledge, and the models
trained on it, above the paradox.

\begin{figure}[t]\revstart
  \centering
  \includegraphics[width=\textwidth]{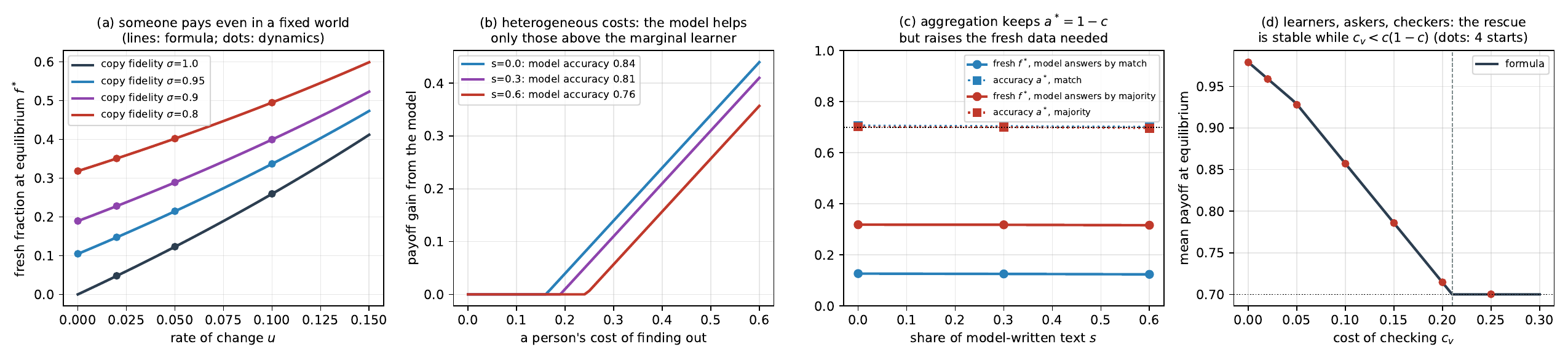}
  \caption{Extensions of the ecosystem model ($u=0.05$, $c=0.3$, $s=0.3$
  unless varied). (a)~Fresh fraction at equilibrium against the rate of
  change, for copying fidelities $\sigma$ (lines: Eq.~\ref{eq:eco}; dots:
  dynamics); with $\sigma<1$ it is positive even in a fixed world.
  (b)~Heterogeneous costs: payoff gain from the model by a person's cost of
  finding out. (c)~A model that states the majority of its corpus keeps
  $a^*=1-c$ but needs more fresh data than one that matches probabilities.
  (d)~Learners, askers, and checkers: mean payoff at equilibrium against the
  cost of checking (line: Proposition~\ref{prop:check}; dots: four starting
  mixtures).}
  \label{fig:eco2}
\end{figure}
\end{revised}

\section{Filters are social learning strategies}
\label{sec:filters}

Most pipelines do not train on raw synthetic data; they filter it. In
social-learning terms, a filter is a rule for choosing what to copy, and
the theory distinguishes three kinds.

A \emph{conformist} filter prefers what is common. Selecting synthetic
samples by their likelihood under the generator, by the perplexity of a
reference model, or by overlap with existing text are conformist in this
sense: the criterion is resemblance, and rare content is by definition less
resembled. Theory predicts that such filters erode variation faster than
copying at random \citep{boyd1985}, and the evidence agrees. In image
models, sampling biased toward quality preserves precision while
accelerating the loss of diversity \citep{alemohammad2024}, and selection
by typicality precipitates collapse, doing worse than random selection
\citep{qiao2026}; iterated curation by any fixed score concentrates the
model on the highest-scoring samples \citep{ferbach2024}. Our text results
are consistent (Table~\ref{tab:static}). For the trigram model, filtering
by the generator's own likelihood is indistinguishable from discarding at
random; for the transformers, whose likelihood ranks whole lines more
sharply, it is clearly worse than random and accelerates the loss of rare
words. In the character transformer, keeping the most likely half of the
sampled lines gives 5.18 against 3.93 bits per character at generation 6
for a random half (Welch $p=0.02$, $d=4.7$, bootstrap 95\% CI of the
difference $[0.85, 1.52]$; Fig.~\ref{fig:neural}); in the pretrained model
it lowers rare-word coverage to 0.05 against 0.57 for a random half (Welch
$p<10^{-8}$, Mann--Whitney $p=0.008$, $d=24$) and the variety of its own
samples to 267 distinct trigrams per 1\,000 words against 940
(Fig.~\ref{fig:opt}). A random half is indistinguishable from no filter in
every model. Filtering by a reference model's likelihood, or by matching
human n-grams, loses nearly every rare word within twelve generations,
degrades perplexity on human text more than doing nothing, and sharply
reduces the variety of the model's own output. A filter that inverts the
bias, favouring rare words, preserves more of them but degrades perplexity
faster: it keeps rare words by keeping the errors around them.

A \emph{payoff-biased} filter keeps what scores best on a task, such as
outputs that a reward model rates highly or that pass a check. It
concentrates the population on the current best and, like selection,
reduces variation even when it improves average quality; iterated
self-improvement narrows output diversity even when it preserves
correctness \citep{briesch2023,ferbach2024}, and verifiers can raise a
model only to the level of the verifier \citep{yi2026}. \rev{We did not
test payoff-biased filters ourselves.}

A \emph{critical} filter tests a copy against the world and replaces it
when it fails. Verification against ground truth \citep{feng2025},
correction toward an expert \citep{gillman2024}, and training on one's own
reasoning only when it reaches a verified answer \citep{zelikman2022}
belong here. \rev{We have not yet tested a critical filter in our own
lineages; the evidence for this class comes from the work cited.}
The distinction that matters is not how much a filter discards but where
its criterion gets its information: from resemblance to what the system
already has, or from the world.

Two results complicate the picture and are worth reconciling. Selecting
synthetic text with a detector of machine-generated text can prevent
collapse \citep{drayson2025}: such a detector penalises the
over-represented features of model output, which makes it closer to an
anti-conformist filter than to the likelihood filters above, and our
rare-favouring filter shows the same partial benefit. And selecting
synthetic data by n-gram similarity to human text had little effect in a
setting that mixes synthetic with human data at the token level
\citep{zhu2025toedit}; in our fully recursive setting the same criterion,
applied every generation, compounds.

\section{Discussion}
\label{sec:discussion}

\begin{revised}
\paragraph{What the reading buys.} The statistical theory of collapse is
more detailed than Eq.~\ref{eq:chain} wherever both apply. What cultural
evolution adds is a separation of two mechanisms that collapse analyses
merge, each with its own signature: lossy transmission removes knowledge
and is predicted by drift on frequencies; staleness replaces it with
outdated values and scales with the rate of change. It also adds a frame in
which the supply of fresh information is a variable, with a price that can
be computed in both regimes, and a taxonomy in which the effect of a filter
can be predicted from where its criterion gets its information.

\paragraph{Predictions.} The analysis makes predictions that larger
experiments could refute: in a fixed world, the coverage a lineage keeps
should follow drift on its frequencies, with departures pointing to
knowledge held outside the chain; in a changing world, models should give
outdated answers in proportion to the rate of change of a domain and to
their reliance on copied text, so that fast-changing domains degrade first
as synthetic text accumulates; lineages that carry their weights forward
should hold more stale facts than lineages that restart from fixed
pretrained weights; likelihood- and typicality-based filters should
accelerate the loss of rare knowledge at scale as they do in small models;
and, in deployed systems, the share of users who verify answers should
govern the long-run reliability of models trained on what users write.

\paragraph{One class or two.} The appendix compares the one-rate chain of
the first version with a chain of two classes of knowledge, fitted to fully
synthetic training (Table~\ref{tab:chainfit}). Two classes describe the
all-synthetic trajectories better, yet predict the fresh-data plateaus of
trigram lineages worse than one class. The drift baseline explains why:
how easily a word is lost depends on how many copies of it a corpus holds,
which fresh data change, so any chain with fixed per-class rates fitted to
one mixture misstates the others. For trigram lineages we therefore rely on
the drift baseline and report the chains only for comparison.

\paragraph{Limitations.} The language models are small, the largest
having 125 million parameters, chosen so that knowledge could be counted
or probed exactly. Rare-word coverage is a corpus measure; we added
model-level measures (rare-word likelihood, factual accuracy and stale
answers) but only for OPT-125m. The fact world is synthetic and simple:
one attribute per person, uniform changes, and facts stated in a fixed set
of templates; real knowledge is correlated and changes unevenly. Several
experiments use three or four seeds; their between-seed variation is small
relative to the effects reported, but rank-based tests at these sizes have
limited power. The ecosystem model treats facts as independent and
strategies as population-level averages; we relaxed equal costs,
probability matching, and the set of strategies, but not the assumption
that the next model is trained on what people write, and we have not tested
it with real users. Our filter experiments cover conformist filters only;
payoff-biased and critical filters are supported here by prior work, not
by our own lineages.
\end{revised}

\paragraph{A wider correspondence.} The identification is one instance of
a broader parallel between the layers of modern training -- pretraining,
fine-tuning, adaptation in context, and training on other models' output --
and the layers of biological adaptation -- evolution, individual learning,
and social learning \citep{stephens1991,boyd1985,henrich2015}. The parallel
is not one of mechanism; it is that both are stacks of adaptive processes
with different speeds and costs that must divide the work of keeping up
with a changing world, and results that depend only on rates, costs, and
channels transfer between them. \rev{The pretrained weights restored at
every generation play the part of a slow layer that holds what the fast
layers keep losing, and the inherited weights the part of individual memory
that keeps what has since become false.}

\section{Conclusion}

\rev{Read through cultural evolution, model collapse is two old problems
met at industrial scale: copies that lose what they copy, and copies of a
world that has moved on.} Cultural evolution's answer was never to stop
copying: it was to keep learners in the population and to check copies
against the world. Taking that answer seriously changes the question
machine learning asks about collapse. The question is not only how much
fresh data a model needs, but who produces it, what it costs them, and
whether the systems built on their work leave them any reason to keep
producing it.

\paragraph{Code and data.} Code for every experiment and figure, and the
per-seed results, are available \coderepo.

\bibliographystyle{tmlr}
\bibliography{references}

\appendix

\section{Experimental details}
\label{app:details}

\paragraph{Word trigram models.} Sentences of each book (Project
Gutenberg; lower-cased, split into sentences at full stops, question and
exclamation marks, tokenised into words and punctuation marks, and kept if
4 to 100 tokens long) are divided at random into a test set (10\%), a
reference set (20\%, used only by the reference-likelihood filter), and a
pool (70\%). Each model interpolates trigram, bigram, and unigram estimates
with weights 0.6, 0.3, and 0.1; the unigram is add-0.01 smoothed. Sentences
are sampled word by word from the interpolated distribution until the end
token; samples shorter than four tokens are discarded. Each generation is
trained on 30\,000 tokens, for 12 generations and 5 seeds. Rare words are
the pool's words outside its 1\,000 most frequent. Filters sample 60\,000
tokens and keep sentences, in order of their score, up to 30\,000 tokens.
\rev{\emph{Fresh data.} At every generation the pool is shuffled with the
seed's random generator and whole sentences are taken in order until the
token budget is met. Draws are therefore without replacement within a
generation and independent across generations: a fresh sentence may have
appeared in an earlier corpus, including generation~0. Repeated draws do
not inflate coverage relative to the fully fresh reference, which is drawn
the same way. In the moving world, fresh sentences are drawn with
replacement from the current window.} \rev{\emph{Neutral drift.} The
synthetic part is drawn as a multinomial sample of $(1-f)N$ tokens from
the previous corpus's token counts; the fresh part is drawn exactly as for
the trained lineages.}

\paragraph{Character transformer.} A 2-layer transformer encoder with
causal masking (width 128, 4 heads, feed-forward width 512, dropout 0.1,
context 128 characters, alphabet of 35 characters), trained from scratch
each generation with AdamW (learning rate $2\times10^{-3}$, batch 32,
1\,000 steps) on 60\,000 characters of Darwin's text; samples at
temperature~1; 6 generations, 3 seeds. Filters sample twice the corpus
and keep half of the lines, by likelihood under the current model or at
random.

\paragraph{OPT-125m.} The pretrained \texttt{facebook/opt-125m} is
fine-tuned afresh from its pretrained weights at every generation, for 2
epochs with AdamW (learning rate $5\times10^{-5}$, batch 32, bfloat16
autocast), on 4\,000 blocks of 64 tokens. The human pool is the
wikitext-2 (raw) training split with headings removed, cut into
consecutive 64-token blocks; perplexity is measured on the first 600
blocks of the test split. Synthetic blocks are sampled from the start
token with pure sampling (temperature 1, no top-$k$ or nucleus
truncation), exactly 64 new tokens each. Fresh blocks are drawn without
replacement within a generation and independently across generations. The
likelihood filter samples 8\,000 blocks and keeps the 4\,000 with the
lowest mean per-token negative log-likelihood under the current model; the
random filter keeps a random 4\,000. Rare words are the pool's words
outside its 1\,000 most frequent; variety is measured on 1\,000 blocks
sampled from each generation's model. 10 generations, 5 seeds.

\begin{revised}
\paragraph{arXiv lineages.} 20\,700 abstracts of arXiv cs.CL papers (the
first 450 listed per month, January 2023 to October 2026), in submission
order, cut into 77\,904 blocks of 64 tokens. The world at generation $t$ is
the window of 6\,000 blocks starting at $t$ times the speed (0 or 6\,000);
every tenth block is held out (600 for evaluation) and fresh blocks are
drawn without replacement from the rest. Training is as for OPT-125m
above; inherit lineages continue from the previous generation's weights
with a new optimizer. Rare and frequent words are defined on the current
window's training blocks. A word's NLL is the sum of the NLL of its tokens
in held-out human blocks (the first word of each block is skipped, as it
may be cut). Excess NLL is the difference to a lineage with fully fresh
data on the same window and seed. 10 generations, 4 seeds.

\paragraph{Fact world.} 2\,000 fictional names of two parts, each of two or
three syllables from a fixed set; 50 city names that are single tokens for
OPT's tokenizer. Human text renders people drawn with probability
proportional to $1/r$ (rank $r$) with one of 16 templates. Probes: ``X
lives in'', ``X moved to'', ``X is based in'', ``X resides in''; the model's
next-token probabilities are renormalised over the 50 cities and averaged
over probes. Training as for OPT-125m but for 3 epochs at learning rate
$2\times10^{-4}$, chosen on a fully fresh generation~0 so that head, mid,
and tail facts span high, middle, and low accuracy (0.99, 0.65, 0.31).
Changes: each fact moves to a uniformly chosen other city with probability
$u$ between generations. Coverage of a fact in a corpus (reported in the
code) is whether some sentence states the person with the current city.
Calibration of Eq.~\ref{eq:chain}: least squares on the static lineage's
accuracy trajectory for each fresh fraction, frequency class, and seed,
$\phi,\sigma\in[0,1]$, starting from the observed generation-0 accuracy.
10 generations, 3 seeds; one lineage takes about 3 minutes on one consumer
GPU.
\end{revised}

\paragraph{Statistics.} The unit of analysis is the seed (one lineage).
We report Welch's $t$-test, the Mann--Whitney $U$ test, Cohen's $d$ with
pooled standard deviation, and a percentile bootstrap 95\% confidence
interval of the difference in means (10\,000 resamples).

\section{One-rate and two-class chains}
\label{app:chain}

\rev{These are the chains of the first version of this paper, kept for
comparison (Section~\ref{sec:discussion}).} Each form is fitted by least
squares on log coverage to the mean relative coverage of the
all-synthetic condition, and applied unchanged to the fresh-data
conditions with $\rho = 1 - f$, using $q_{t+1}=(1-\rho)+\rho(1-u)q_t$. The
two-class form is $q_t = w\,q_t(u_1) + (1-w)\,q_t(u_2)$. To choose between
forms without using any fresh-data run, each is also fitted to the first
half of the all-synthetic generations and scored on the second half
(held-out error, maximum absolute error). The prediction errors are the
largest absolute differences between predicted and observed coverage over
the fresh-data conditions, at the final generation and over all
generations.

\begin{table}[h]
  \caption{One-class and two-class chains of the first version, fitted to
  fully synthetic training only. $^{*}$: the form preferred by the held-out
  generations of the all-synthetic condition. \rev{For trigram lineages the
  zero-parameter drift baseline has prediction errors of at most 0.007
  (final) and 0.028 (any generation).}}
  \label{tab:chainfit}
  \centering
  \small
  \begin{tabular}{llllll}
  \toprule
  & & & held-out & \multicolumn{2}{c}{prediction error} \\
  \cmidrule(lr){5-6}
  setting & form & parameters & error & final & any generation \\
  \midrule
trigram, Darwin & one & $u=0.15$ & 0.115 & 0.032 & 0.107 \\
 & two$^{*}$ & $w=0.49$, $u_1=0.46$, $u_2=0.08$ & 0.029 & 0.081 & 0.091 \\
\addlinespace
trigram, Melville & one & $u=0.16$ & 0.111 & 0.029 & 0.120 \\
 & two$^{*}$ & $w=0.55$, $u_1=0.45$, $u_2=0.08$ & 0.035 & 0.082 & 0.092 \\
\addlinespace
character transformer & one & $u=0.47$ & 0.049 & 0.072 & 0.230 \\
 & two$^{*}$ & $w=0.83$, $u_1=0.97$, $u_2=0.20$ & 0.002 & 0.037 & 0.054 \\
\addlinespace
OPT-125m & one & $u=0.07$ & 0.253 & 0.058 & 0.159 \\
 & two$^{*}$ & $w=0.28$, $u_1=0.82$, $u_2=0.02$ & 0.039 & 0.030 & 0.044 \\
  \bottomrule
  \end{tabular}
\end{table}

\begin{revised}
\section{Derivations for the ecosystem model}
\label{app:proofs}

\paragraph{Coverage correction in the simulation.} In the agent-based
simulation each of $N$ people writes about one fact drawn uniformly from
$K$. A fact receives at least one human-written statement in a generation
with probability $\kappa=1-(1-1/K)^N$; a fact that receives none keeps the
previous model's value. Of the facts that do receive one, a share $1-s$
enter the corpus from people, and the last writer is a learner with
probability $h$. The fresh fraction per fact is therefore $f=(1-s)\kappa
h$, and Proposition~\ref{prop:eco} gives $h^* = f^*/((1-s)\kappa)$. For
$N=K=4\,000$, $\kappa=0.632$.

\paragraph{Proof of Proposition~\ref{prop:check}.} Payoffs are $\pi_L=1-c$,
$\pi_A=a$, and $\pi_C=1-c_v-(1-a)c$. Checkers and learners both write the
truth, so the fresh fraction is $f=(1-s)(x_L+x_C)$, and accuracy follows
Proposition~\ref{prop:eco}'s recursion, increasing in $f$. Askers and
checkers earn the same when $a(1-c)=1-c-c_v$, that is at
$a^{**}=1-c_v/(1-c)$. At $a^{**}$, $\pi_C-\pi_L=a^{**}c-c_v$, which is
positive whenever $c_v<c(1-c)$ (then $a^{**}>1-c>c_v/c$ for $c<1$), so
learners cannot invade; mean payoff is $\pi_A=\pi_C=a^{**}$. Along the
asker-checker edge, $\pi_C-\pi_A=(1-a)(1-c)-c_v$ decreases in $a$, and $a$
increases with the share of checkers, so a perturbation toward more
checkers raises $a$ and makes checking pay less, and vice versa: the
equilibrium is stable. If even $x_C=1$ cannot reach $a^{**}$, that is if
$a_{\max}=(1-u)(1-s)/(1-(1-u)s\sigma)<a^{**}$, checkers always earn more
than askers, everyone checks, and mean payoff is $1-c_v-(1-a_{\max})c$.
If $c_v\ge c(1-c)$, then at $a=1-c$ checkers earn no more than learners and
the population returns to the learner-asker equilibrium of
Proposition~\ref{prop:eco}.

\paragraph{Numerical checks.} Mean-field dynamics of $a$ and the strategy
shares, with replicator updates (step 0.5) and mutation $10^{-4}$, run for
6\,000 generations, match $f^*$ of Proposition~\ref{prop:eco} to three
decimals for $\sigma\in\{1, 0.95, 0.9, 0.8\}$ and $u\in\{0, 0.02, 0.05,
0.1\}$, and $\bar W^{**}$ of Proposition~\ref{prop:check} to within 0.001
from four starting mixtures for $c_v$ from 0 to 0.25. Heterogeneous costs
use 61 cost groups uniform on $[0, 0.6]$, each with its own learner share.
The aggregation check simulates 4\,000 facts, each with nine statements
per corpus; after a change all statements of the fact are false; the model
states either a random statement (probability matching) or the majority.
\end{revised}

\end{document}